\documentclass[letterpaper, 10 pt, conference]{ieeeconf}  

\IEEEoverridecommandlockouts                              

\usepackage{amsmath}
\usepackage{amssymb}

\usepackage{times}
\usepackage{epsfig}
\usepackage{graphicx}
\usepackage{amsmath}
\usepackage{mathabx}
\usepackage{bm}
\usepackage{amssymb}

\usepackage{tabularx}
\usepackage{amsmath}
\usepackage{siunitx}
\usepackage{booktabs}
\usepackage{multirow}
\usepackage[table,xcdraw]{xcolor}
\usepackage{float}
\usepackage[normalem]{ulem}
\useunder{\uline}{\ul}{}

\usepackage{cite}
\usepackage{booktabs}
\usepackage{caption}
\usepackage{subcaption}
\usepackage{hyperref}
\usepackage{algorithmic}
\usepackage[ruled,vlined, linesnumbered]{algorithm2e}

\makeatletter
\newdimen\commentwd
\let\oldtcp\tcp
\def\tcp*[#1]#2{
\setbox0\hbox{\textcolor{blue}{#2}}%
\ifdim\wd\z@>\commentwd\global\commentwd\wd\z@\fi
\oldtcp*[r]{\leavevmode\hbox to \commentwd{\box0\hfill}}}

\title{\LARGE \bf Image Masking for Robust Self-Supervised Monocular Depth Estimation}

\author{Hemang Chawla$^1$, Kishaan Jeeveswaran$^1$, Elahe Arani$^{1,2}$*, and Bahram Zonooz$^{1,2}$*
\thanks{Authors are with $^1$Advanced Research Lab, NavInfo Europe, The Netherlands, and $^2$Department of Mathematics and Computer Science, Eindhoven University of Technology, The Netherlands.\newline
Contact: {\tt\small hemang.chawla@navinfo.eu}\newline
*Contributed equally.\newline
Code: \href{https://github.com/NeurAI-Lab/MIMDepth}{https://github.com/NeurAI-Lab/MIMDepth}}%
}

\begin{document}

\maketitle
\thispagestyle{empty}
\pagestyle{empty}

\begin{abstract}
Self-supervised monocular depth estimation is a salient task for 3D scene understanding. Learned jointly with monocular ego-motion estimation, several methods have been proposed to predict accurate pixel-wise depth without using labeled data. Nevertheless, these methods focus on improving performance under ideal conditions without natural or digital corruptions. The general absence of occlusions is assumed even for object-specific depth estimation. These methods are also vulnerable to adversarial attacks, which is a pertinent concern for their reliable deployment in robots and autonomous driving systems.  We propose MIMDepth, a method that adapts masked image modeling (MIM) for self-supervised monocular depth estimation. While MIM has been used to learn generalizable features during pre-training, we show how it could be adapted for direct training of monocular depth estimation. Our experiments show that MIMDepth is more robust to noise, blur, weather conditions, digital artifacts, occlusions, as well as untargeted and targeted adversarial attacks. 
\end{abstract}
\section{Introduction}
\label{sec:introduction}
\begin{figure*}[htbp]
\centering
  \includegraphics[width=0.88\textwidth, trim=0 0 3.5cm 0, clip]{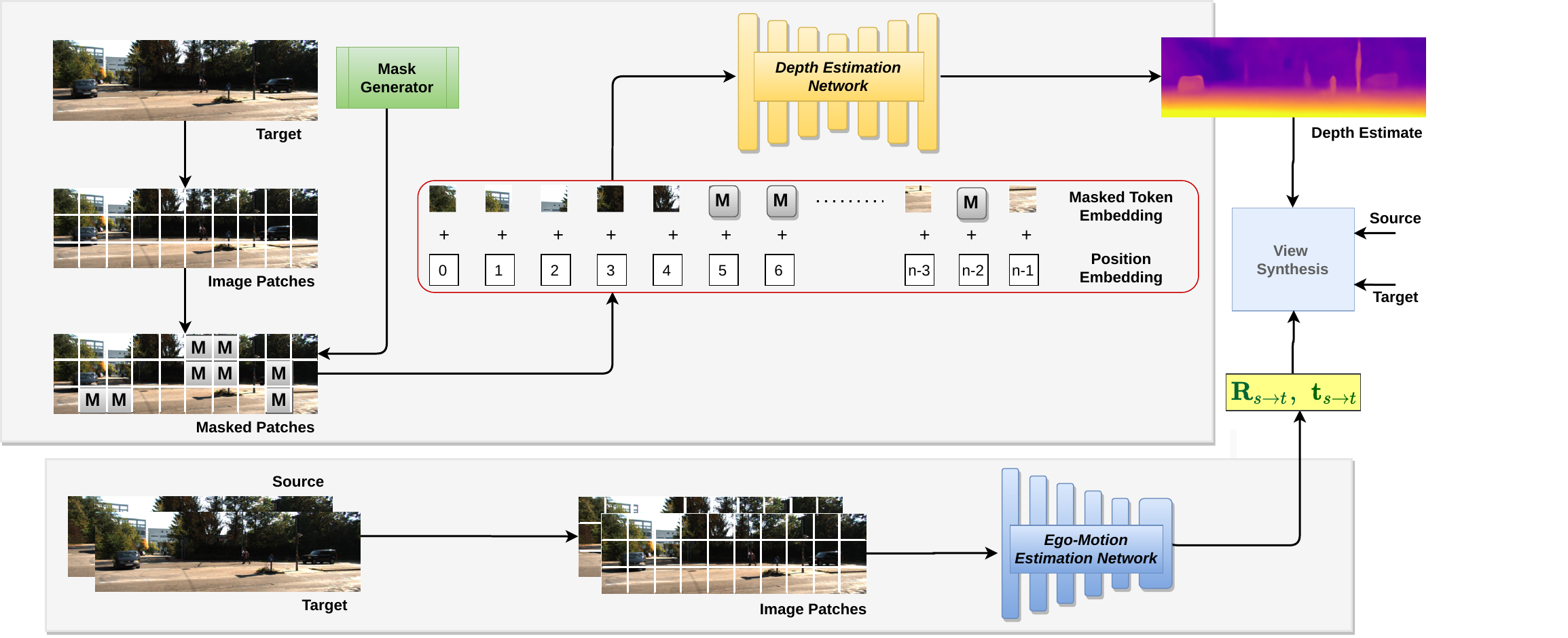}
  \caption{An overview of 
  Masked Image Modeling for Depth Estimation. The method learns to predict depth from the masked as well as unmasked patches with a better understanding of the global context.} 
\label{fig:framework}
\end{figure*}

Depth estimation is an essential component of vision systems that capture 3D scene structures for applications in mobile robots, self-driving cars, and augmented reality.  Although expensive and power-hungry LiDARs offer high-accuracy depth measurements, the ubiquity of low-cost, energy-efficient cameras makes monocular depth estimation techniques a popular alternative. Traditional methods often estimate depth from multiple views of the scene~\cite{schonberger2016structure}. Instead, deep learning methods have demonstrated depth estimation from a single image. Nevertheless, supervised depth estimation approaches~\cite{lee1907big, yin2019enforcing} require ground truth labels, making it difficult to scale.
On the contrary, self-supervised depth estimation approaches are trained without ground truth labels by using concepts from traditional structure-from-motion and offer the possibility of training on a wide variety of data~\cite{chawlavarma2021multimodal, guizilini20203d}. However, the deployment requires a focus on the generalizability and robustness of models beyond performance under ideal conditions~\cite{chawlavarma2022adversarial}.

Recently, MT-SfMLearner~\cite{visapp22} showed that the transformer architecture for self-supervised depth estimation results in higher robustness to image corruptions as well as against adversarial attacks.  This is attributed to transformers utilizing the global context of the scene for predictions, unlike convolutional neural networks that have a limited receptive field. However, most research in self-supervised monocular depth estimation focuses primarily on achieving excellent performance on the independent and identically distributed (i.i.d.) test set.  It is assumed that the images are free from noise (e.g. Gaussian) and blur (e.g. due to ego-motion or moving objects in the scene), have clear daylight weather, and are without digital artifacts (e.g. pixelation).  Even for the task of object-specific depth estimation~\cite{lee2021realtime}, it is assumed that the objects are without occlusions. Finally, the robustness of methods against adversarial attacks is not considered, which is a pertinent concern for safety while deploying deep learning models.  

Since the performance and robustness of the models are determined by the learned representations, influencing the encoding of features could lead to more robust estimations. We hypothesize that integrating Masked Image Modeling (MIM) into the training of depth estimation would result in learning features that make the model more robust to natural and digital corruptions as well as against adversarial attacks, by modeling the global context in a better way. MIM is a technique that has been used until now for self-supervised representation learning in pre-training of image transformers~\cite{bao2021beit, xie2022simmim,he2022masked,wei2022masked}.
MIM pre-training involves masking a portion of image patches and then using the unmasked portion to predict the masked input~\cite{xie2022simmim, he2022masked} or its features~\cite{bao2021beit, wei2022masked}. It models long-range dependencies and focuses on the low-frequency details present in the images.
However, when pre-trained models are fine-tuned for downstream training, the general features that were learned could possibly be overwritten.  Instead, adapting MIM for direct training of a task, such as depth estimation, could lead to richer learned representations that make the model more robust and generalizable. 

While both MIM and depth estimation are self-supervised methods, they differ in how they are trained. MIM, used in pre-training of a network, learns by reconstructing the input image generally passed through an autoencoder. Instead, the self-supervised depth estimation network is trained along with a self-supervised ego-motion estimation network, whose output is used to synthesize adjacent images in the training set via the perspective projection transform~\cite{zhou2017unsupervised}.  
Thus, applying MIM to depth estimation requires different considerations than its use for pre-training. Here, we examine the following questions:
\begin{itemize}
    \item Would integrating MIM to the depth and/or ego-motion estimation networks result in improved robustness of depth estimation?
    \item  MIM has been shown to work well with either blockwise masking~\cite{bao2021beit}, or random masking with high mask size~\cite{xie2022simmim}. Which masking strategy would work better for depth estimation?
    \item MIM pre-training uses a relatively high mask ratio and mask size~\cite{bao2021beit, xie2022simmim, he2022masked} due to more information redundancy in images than in sentences. Would the high mask ratio or high mask size used for MIM pre-training be suitable for integrating it into depth estimation? 
    \item MIM has been shown to result in better features for downstream tasks when its loss is applied only to masked regions~\cite{xie2022simmim}. Would similarly applying the loss on only the masked regions result in a more robust depth estimation?
\end{itemize}

With our proposed method MIMDepth, we demonstrate that applying blockwise masking with a relatively lower mask ratio (than MIM pre-training) only to the depth estimation network, with a loss on the complete image, results in improved robustness to natural and digital corruptions, occlusions, as well as untargeted and targeted adversarial attacks.
It is additionally found to improve the performance of the ego-motion estimation network, while maintaining a competitive performance of the depth estimation network on the i.i.d. test sets.

\section{Related Works}


\textbf{Self-supervised Monocular Depth Estimation}
One of the challenging tasks of interest in 3D scene understanding is monocular depth estimation. Although self-supervised depth estimation was introduced for stereo pairs~\cite{godard2017unsupervised}, it was soon extended to a monocular setup~\cite{zhou2017unsupervised}. 
Monocular self-supervised approaches to depth estimation have the advantage of not requiring any labels and can learn from a wide variety of data from multiple sources. 
Over the years, improvements have been made to deal with challenges due to occlusions~\cite{godard2019digging}, dynamic objects~\cite{gordon2019depth, klingner2020selfsupervised, ranjan2019competitive}, and scale-consistency issues~\cite{bian2019unsupervised,chawlavarma2021multimodal} and more. While most methods generally used 2D convolutional architectures, a 3D convolutional architecture was proposed to estimate depth from symmetrical packing and unpacking blocks that preserve depth details~\cite{guizilini20203d} . Recently, MTSfMLearner~\cite{visapp22} has shown that transformers can also be used for depth and pose estimation resulting in comparable performance, but improved robustness to natural corruptions and adversarial attacks due to their global receptive fields. Although other methods that use transformers have also been proposed~\cite{guizilini2022multi, yang2022depth}, they do not consider the robustness of their proposed approaches.  We show that integrating mask image modeling trains networks to identify long-range dependencies and could further improve the robustness of depth estimation. 

\textbf{Masked Image Modeling}
Masked image modeling is a method for self-supervised representation learning through images corrupted by masking. This was developed following masked language modeling (MLM)~\cite{kenton2019bert}. These methods are based on replacing a portion of the tokenized input sequence with learnable mask tokens and learning to predict the missing context using only the visible context.  iGPT~\cite{chen2020generative} operates on clustered pixel tokens and predicts unknown pixels directly. ViT~\cite{dosovitskiy2020vit} explores masked patch prediction by predicting the mean color.  On the contrary, BEIT~\cite{bao2021beit} operates on image patches but uses an additional discrete Variational AutoEncoder (dVAE) tokenizer~\cite{ramesh2021zero} to learn to predict discrete tokens corresponding to masked portions.  BEIT uses special blockwise masking that mitigates the wastage of modeling capabilities on short-range dependencies and high-frequency details similar to BERT. Instead, SiMMIM~\cite{xie2022simmim} and MAE~\cite{he2022masked} show that even random masking with a higher mask ratio or mask size can similarly perform well for self-supervised pretraining from image data. 
However, MIM has not been explored for directly training a relevant task of interest rather than as a pre-training method. We demonstrate how it can be adapted to improve the robustness of self-supervised monocular depth estimation. 

\section{Method}
\label{sec:method}



We propose MIMDepth, a method for masked self-supervised monocular depth estimation (see Figure \ref{fig:framework}) that integrates masked image modeling (MIM) into self-supervised monocular depth estimation. MIMDepth trains the depth estimation network by masking a portion of the input,  while the network learns to predict complete pixelwise depth from the partial observations, including for the masked portions. Hereafter, we elaborate upon the components of the method.

\subsection{Self-supervised Monocular Depth Estimation} We use MT-SfMLearner~\cite{visapp22} as the baseline monocular depth estimation method, which we briefly summarize here. 
Similar to other self-supervised methods, it trains depth and ego-motion estimation networks simultaneously. The training input to the networks is a set of consecutive image triplets $\{I_{-1}, I_0, I_1\}$ from a video. The depth network is trained to output pixel-wise depth for the target image $I_0$, while the ego-motion estimation network is trained to output relative translation and rotation between the image pairs $\{I_{-1}, I_0\}$ and $\{I_0, I_1\}$.  Note that for the ego-motion estimation network, the input pair of images is concatenated along the channel dimension. Both the depth and ego-motion estimation networks utilize a transformer-based encoder DeiT-Base~\cite{touvron2021training} which processes the tokenized inputs at constant resolution. 

The predicted depth and ego-motion are linked together via the perspective projection transform for a given set of camera intrinsics $K$, through which the source images $I_{-1}$ and $I_1$ are warped to synthesize the target image $\hat{I}_0$.  Both the depth and the ego-motion estimation networks are trained simultaneously using a common loss $\mathcal{L}_{depth}$ between $\hat{I}_0$ and $I_0$, which is a weighted sum of two losses. 
First is the appearance-based photometric loss between the original and the synthesized target image, $\mathcal{L}_{photo}$. The photometric loss is itself composed of a structural similarity $(\mathcal{L}_\text{SSIM})$ loss~\cite{wang2004image} and an $\ell_1$ distance loss between the images.  Second is a smoothness loss on the depth output, $\mathcal{L}_\text{smooth}$, for regularization in low-texture portions of the scene. Therefore, 
\begin{equation}
    \mathcal{L}_{depth} = \lambda_1 \cdot \mathcal{L}_\text{SSIM} + \lambda_2 \cdot \ell_1 + \lambda_3 \cdot \mathcal{L}_\text{smooth},
\end{equation}
where $\lambda_1$, $\lambda_2$, and $\lambda_3$ are the hyperparameter weights for the losses.

\subsection{Integrating Masked Image Modeling}
Our method MIMDepth, integrates MIM with self-supervised monocular depth estimation for masked self-supervised depth estimation. While MIM in pre-training applies masking for the network that reconstructs the image, we could mask the input to the depth or the ego-motion estimation networks. 
The input to the depth estimation network is a single image, which is converted into non-overlapping patches with token and position embeddings. 
For integrating MIM with the depth network, we use a mask generator to sample the patch indexes to be masked. The method for mask generation is described in Algorithm~\ref{alg:blockwise_masking_depth}. 
A block with a random aspect ratio is repeatedly selected to be masked in the image until the required mask ratio is achieved.  The indexes to be masked are replaced with learnable masked token embeddings, which are initialized as Gaussian noise. 

We use a blockwise masking strategy similar to BEIT~\cite{bao2021beit}, but we do not use any dVAE to generate targets. We use a mask ratio of 25\% which is higher than that used in MLM but less than that used in MIM pre-training. This is due to the lower information redundancy in the depth of objects than in their texture. As an example, a gray color car would have similar texture across the pixels, but its depth would vary depending upon how it is oriented with respect to the camera. The network is trained to directly predict the depth for the complete image from the partially masked input. However, we do not apply masking to the ego-motion network, as it does not improve depth estimation (see Section~\ref{sec:experiments}). 

SimMIM~\cite{xie2022simmim} shows that it is better to learn to predict over the masked patches rather than the complete image in order to pre-train representations that assist in downstream tasks.   This is because training the network on the complete image would waste learning capability on the reconstruction quality of unmasked portions. However, with MIMDepth, the objective is to directly learn pixel-wise dense prediction including the unmasked portions. Hence, we apply the loss $\mathcal{L}_{depth}$ to the complete target image. 

\begin{algorithm}[t]
\footnotesize
\DontPrintSemicolon
\SetKwInOut{Input}{input}\SetKwInOut{Output}{output}
\SetKwFunction{Random}{random}
\SetKwFunction{Gaussian}{$\mathcal{N}$}
\SetKwFunction{repeat}{repeat}

\SetKwRepeat{Do}{do}{while}

\SetKwData{h}{h}
\SetKwData{w}{w}
\SetKwData{aspect}{aspect}
\SetKwData{MaskRatio}{mask\_ratio}
\SetKwData{MaskSize}{mask\_size}
\SetKwData{NumPatches}{num\_patches}
\SetKwData{MaskRatio}{mask\_ratio}

\Input{unmasked tokens $x_u$ \\
mask size $m_s$ \\
mask ratio $m_r$ \\
image shape ($h \times w$) \\ 
block aspect ratio $a$
} 

\Output{masked tokens $x_m$}
\BlankLine

$n \leftarrow (h \times w) / m_s^2$  \tcp*[0]{num maskable patches} 
$M \leftarrow$ [] \tcp*[0]{initialize mask}
\Do{$|M| < m_r \times n$}{
$s \leftarrow \Random(m_s, m_r \times n - |M|$) \tcp*[0]{select block size}
$r \leftarrow \Random(a, a^{-1}),  a < 1$ \tcp*[0]{select block aspect ratio} 
$x \leftarrow \Random(0, w - \sqrt{s/r}$) ; $y \leftarrow \Random(0, h - \sqrt{s \cdot r}$) \\
$M_{block} \leftarrow \{ (i,j): i \in [y, y+\sqrt{s \cdot r} ], j \in [x, x+\sqrt{s/r}]  $\} \\
$M \leftarrow M \cup M_{block}$ \tcp*[0]{add block to mask}
}
$M \leftarrow M.\repeat(\lfloor h/m_s \rfloor, \lfloor w/m_s \rfloor$) \tcp*[0]{repeat mask to match the transformer patch size}
$x_m \leftarrow x_u \times (1-M)  + \Gaussian(0, \delta) \times M$ \tcp*[0]{generate masked tokens}
\KwRet{$x_m$}
\caption{Blockwise Masking for Depth Estimation Network}
\label{alg:blockwise_masking_depth}
\end{algorithm}

\section{Results}
\label{sec:experiments}
Here, we perform a comparative analysis between the proposed method MIMDepth and existing self-supervised monocular depth estimation methods in terms of their performance on i.i.d. test set, as well as robustness to natural and digital corruptions, occluded views of the scene, and against adversarial attacks. Thereafter, through an ablation study, we resolve the questions posed in Section~\ref{sec:introduction} regarding the mask sampling strategy, size and ratio, as well as the training target for integrating MIM into depth estimation. Finally, we show how integrating MIM into self-supervised monocular depth estimation also results in improved ego-motion estimation on the i.i.d. test set.

\subsection{Settings}
We train the networks on a TeslaV100 GPU for
20 epochs with AdamW optimizer~\cite{loshchilov2017decoupled} at a resolution of 640 × 192 with batchsize 12. The depth and ego-motion encoders are initialized with ImageNet~\cite{deng2009imagenet} pre-trained weights. The learning rate is set at $1e^{-5}$ and decays by a factor of 10 after 15 epochs. 
Unless otherwise stated, the networks are trained on the Eigen-Zhou split~\cite{zhou2017unsupervised} of the KITTI dataset~\cite{geiger2013vision} with 39,810 training, 4424 validation, and 697 test images, respectively. We set the mask size $m_s$, mask ratio $m_r$, aspect ratio $a$ for Algorithm \ref{alg:blockwise_masking_depth} to $16$, $25\%$, and $0.3$, respectively. 

\subsection{Robustness}
While multiple works focus on the performance of depth estimation models, deployment in the real world necessitates models that not only perform well in the ideal conditions, but are also robust to various corruptions as well as adversarial attacks.  

To evaluate the robustness, we study the impact
of natural corruptions and adversarial attacks on depth performance against other methods, namely Monodepth2~\cite{godard2019digging}, Packnet~\cite{guizilini20203d}, and MT-SfMLearner~\cite{visapp22} which use 2D convolutions, 3D covolutions, and transformer-based architectures, respectively. We report the robustness of MIMDepth through the average RMSE for models trained with three different seeds.

\subsubsection{Robustness to Natural Corruptions}
\begin{figure*}[htbp]
\centering
    \includegraphics[width=0.87\textwidth]{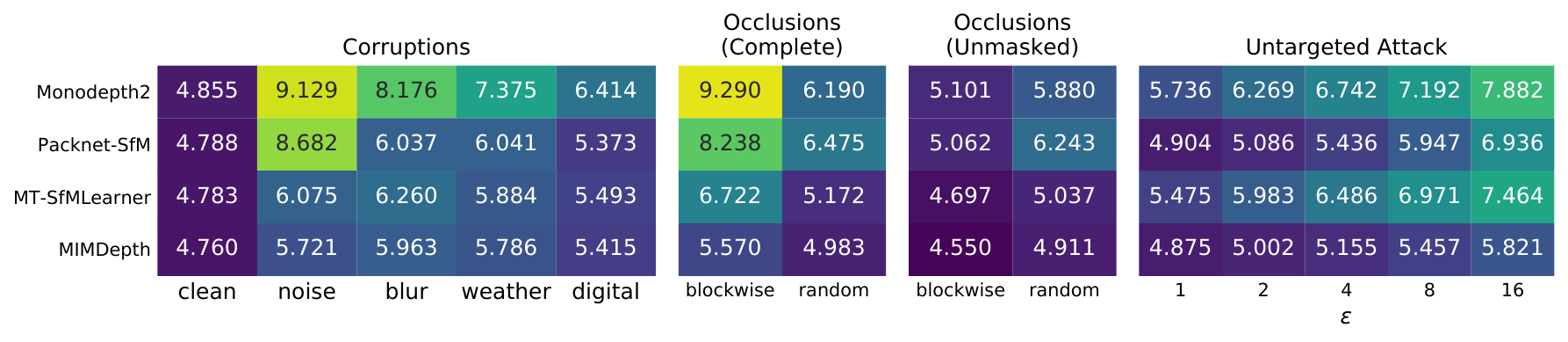}
  \caption{Robustness to natural corruptions, occlusions, and untargeted adversarial attack. Values denote RMSE (m). } 
\label{fig:robust_corruptions}
\end{figure*}

The camera used in an autonomous driving vehicle or a driver assistance system can be subject to different corruptions. Particularly, these corruptions can be due to noise (Gaussian, shot, impulse), or blur (defocus, glass, motion, zoom), different weather conditions (snow, frost, fog, brightness) and even digital aberrations (contrast, elastic, pixelation, JPEG). 
Figure \ref{fig:robust_corruptions} compares the different methods on the four types of corruptions, as well as on clean uncorrupted data.
Note that the proposed MIMDepth outperforms other methods on clean, noisy, blur, and weather corruptions. 
However, PackNet-SfM is better on digital corruption due to its better performance on pixelated corruption, attributed to its architecture's property of preserving depth details through 3D convolutions. Nevertheless, MIMDepth outperforms it on contrast, elastic, and JPEG digital corruptions. 
MIMDepth's robustness can be attributed to the higher level of long-range semantic context that is learned by introducing MIM. The impact of a few corruptions on depth estimates is visualized in Figure~\ref{fig:depth}.


\subsubsection{Occlusion Corruptions}
Monocular depth estimation training requires that the target image is reconstructed using adjacent images. However, due to static as well as dynamic objects, certain regions visible in the target image are occluded in the source image (and vice versa) and cannot be reconstructed. Occlusions are also an important consideration in object-specific depth estimation~\cite{lee2021realtime}. 
However, the role of occlusions has not been studied directly during the test time. Here, we evaluate the impact of occluded context in estimating the depth of objects in the scene. 

\begin{figure*}[t]
\centering
    \includegraphics[width=\textwidth]{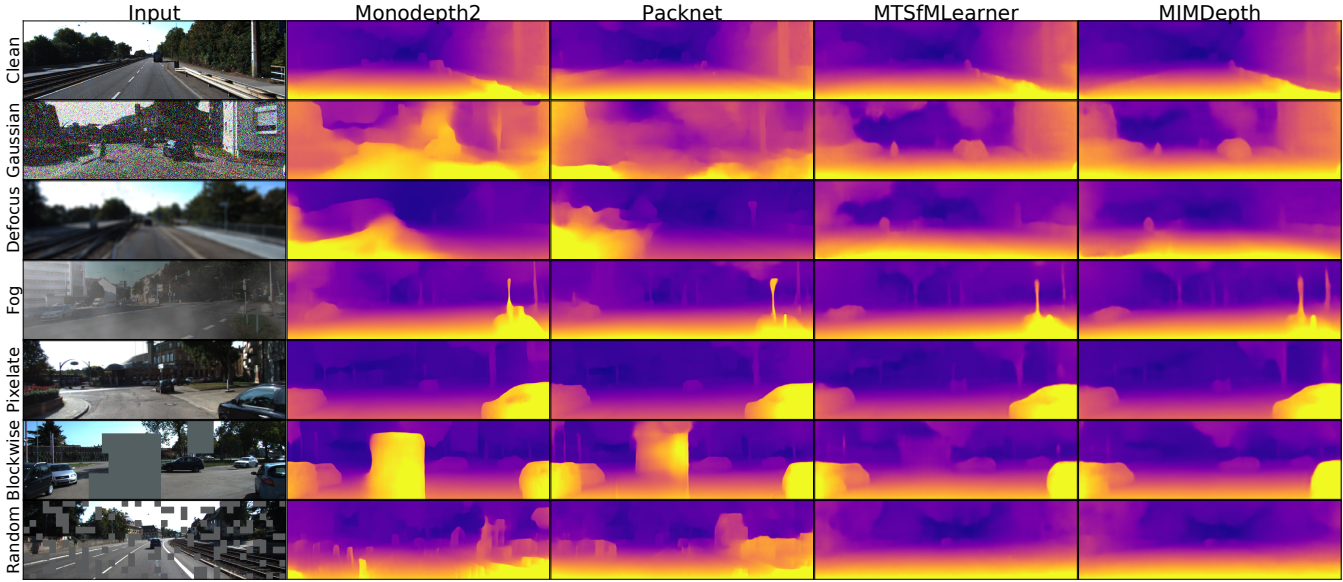}
  \caption{Examples of depth estimation with different corruptions of the scene.} 
\label{fig:depth}
\end{figure*}

Consequently, we simulate occlusions through modified test sets in which portions of the scene are masked out. Pixels in the occluded region are replaced by the average pixel value of the complete image. In particular, we generate occlusions in the images using blockwise and random masking at the 25\% mask ratio. 

The aim is to evaluate the performance of the depth networks on the unoccluded regions of the scene, as well as to study the direct impact of masked depth estimation in predicting the depth on the complete image. This can also be thought to correspond to scenarios in which only a portion of the target object is visible for object-specific depth estimation. 
Figure ~\ref{fig:robust_corruptions} compares the different methods on the occluded test sets. Occlusions are found to impact depth estimates not only in the masked regions but also in the unmasked regions. We observe that MIMDepth is more robust to occlusions than other methods. By using the global context in a better way, MIMDepth is able to have the best predictions on the unmasked regions and is also able to infer plausible depth for the masked regions. The impact of occlusions on depth estimates for an example image is visualized in Figure~\ref{fig:depth}.


\subsection{Adversarial Attacks}
Adversarial attacks are used to fool networks into making an incorrect prediction through imperceptible changes in input images. These also help to measure the generalizability of the network as the data distribution is altered. We consider two types of adversarial attacks.
For the untargeted attack, we generate adversarial examples following \cite{kurakin2016adversarial} at attack strengths of $\epsilon=\{ 1.0,2.0, 4.0, 8.0, 16.0\}$. The gradients are computed with respect to the training loss $\ell_{depth}$ (see Section \ref{sec:method}) and the perturbations are accumulated over $\min(\epsilon + 4, \lceil1.25\cdot\epsilon\rceil)$ iterations. 
For the targeted attack, we generate adversarial examples that are intended to fool networks into predicting flipped depth estimates~\cite{wong2020targeted}. This attack uses the gradients on the RMSE values corresponding to when the target depths are horizontal or vertical flips of the original prediction. Similar to the PGD attack above, we generate adversarial examples with attack strengths of $\epsilon=\{1.0,2.0, 4.0\}$.



\begin{figure*}[htbp]
\centering
    \includegraphics[width=0.95\textwidth]{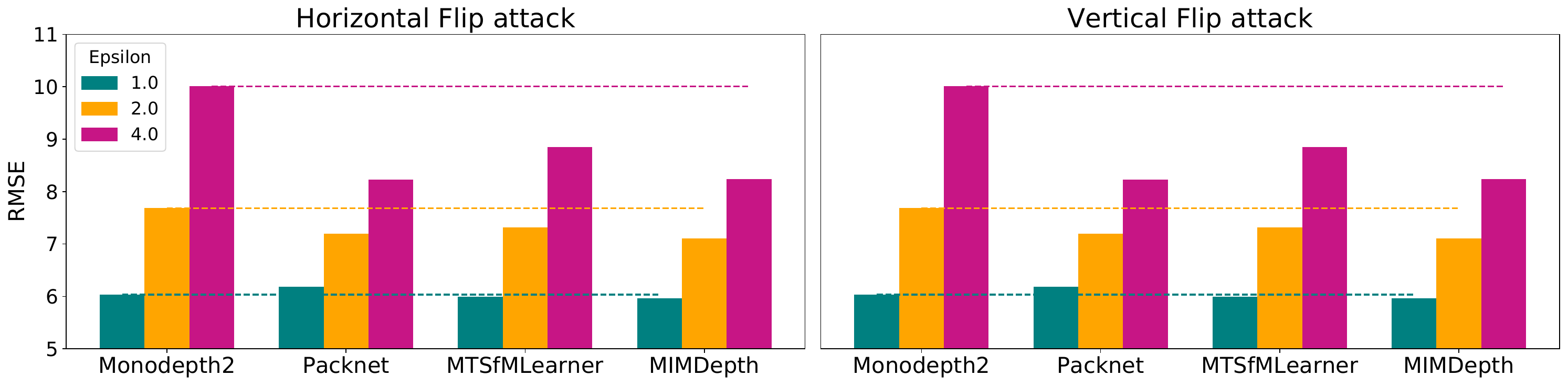}
  \caption{Targeted horizontal and vertical flip adversarial attacks.} 
\label{fig:targeted}
\end{figure*}

The results of the untargeted adversarial attack are shown in Figure~\ref{fig:robust_corruptions}. The results of horizontal and vertical flipped adversarial attacks on depth estimation are shown in Figure~\ref{fig:targeted}. We note that MIMDepth is the most robust to untargeted and targeted attacks. This can be attributed to the better global context that is learned by adapting MIM to depth estimation. 

\subsection{Performance}
\begin{figure}[htbp]
\centering
    \includegraphics[width=0.9\linewidth]{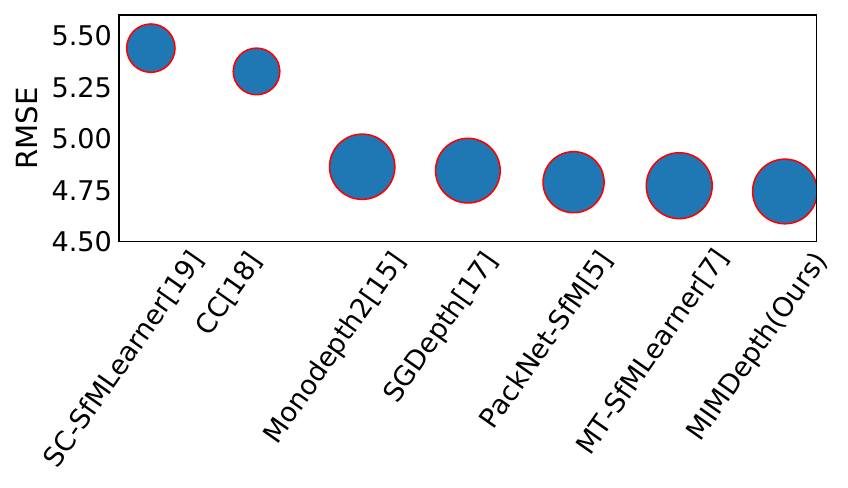}
  \caption{Performance of state-of-the-art self-supervised monocular depth estimation methods. The size of the bubble correlates with the accuracy ($\delta<1.25$) for each method (bigger is better).} 
\label{fig:sota}
\end{figure}

It is noted that the accuracy of the model is often at odds with the robustness~\cite{tsipras2018robustness}. Hence, we also compare MIMDepth with other state-of-the-art methods for its performance on the i.i.d. test set in Figure \ref{fig:sota}. For a fair comparison, we consider methods that predict depth only from single images, do not use additional supervised semantic information and train only on KITTI. MIMDepth is able to achieve performance comparable to that of other state-of-the-art methods on both error (RMSE) and accuracy ($\delta < 1.25)$ metrics~\cite{eigen2014depth}. Overall, this shows that integrating MIM into monocular depth estimation improves the robustness of depth estimation while maintaining comparable performance on the i.i.d. test set.

\textit{Model Size and Computational Efficiency:} The expansion in terms of depth model size is limited to the mask token with dimension \emph{d=768} parameters, which is insignificant compared to the baseline depth model size (183 M). Thus, the memory overhead is marginal. Moreover, since image masking is not used during prediction, the inference speed remains the same.  

\subsection{Ablation Studies}
Here, we resolve the questions posed in Section~\ref{sec:introduction} about the integration of MIM into self-supervised monocular depth estimation through ablation over different configurations. We use the RMSE on clean input and robustness to occlusion corruptions (for complete image) as the metrics for comparison. 

\label{section:results_mask_sampling_strategy}
MIMDepth applies blockwise masking only on the depth network. We have shown that masking the depth network improves the robustness  of the model.
Here, we study the utility of applying MIM also on the ego-motion network. We also compare the use of blockwise and random mask sampling strategies for training the networks.
From Table \ref{tab:ablation_strategy_obstacles }, we observe that blockwise masking helps the depth estimation network to reduce the mean RMSE over clean and corrupted input. Random masking does not perform as well. This is because block-wise masking is better at inhibiting short-range connections to prevent the network from impainting depth. Furthermore, block-wise masking is more reflective of how specific objects may be occluded rather than random occlusions across the scene. 
We also note that masking the ego-motion network along with the depth network does not improve depth estimation performance on clean data or robustness to occlusions. This could be attributed to the ego-motion network learning the translation and rotation from the shared context of the image pair, which is reduced by masking and is ineffective in improving the dense depth predictions. 

\begin{table}[b]
\centering
\begin{tabular}{|c|c|c|c|c|}
\hline
\multicolumn{1}{|c|}{\textbf{Depth}} & \multicolumn{1}{c|}{\textbf{Ego-Motion}} & \multicolumn{1}{l|}{\textbf{Clean}} & \multicolumn{1}{l|}{\textbf{Blockwise}} & \multicolumn{1}{l|}{\textbf{Random}}   \\ \hline \hline
B & -& 4.760 & 5.570 & 4.983  \\ 
R & - & 4.777 & 6.592 & 5.071  \\
B & B & 4.788  & 5.614 & 5.009   \\
B & R & 4.789 & 5.579 & 5.010    \\
\hline
\end{tabular}
\caption{Comparing impact of masking strategies for depth and ego-motion networks. B refers to blockwise masking and R refers to random masking. Values denote RMSE (m).}
\label{tab:ablation_strategy_obstacles }
\end{table}


MIMDepth uses a mask ratio and size of 25\% and 16 respectively. We also compare with the use of a larger mask size or mask ratio in Table \ref{tab:ablation_size_ratio_loss_obstacles}. 
We observe that mask size 16 and mask ratio of 25\% have the lowest mean RMSE over clean and corrupted input. Note that this value is lower than the masking ratio of 40\% used by BEIT~\cite{bao2021beit} or the mask size of 32 used by SimMIM~\cite{xie2022simmim} for pre-training. This can be attributed to the lower information redundancy for depth than that for image reconstruction (e.g. a vehicle with low texture variation and color consistency could have varying relative depth across its body, depending upon its orientation). 

\begin{table}[b]
\centering
\begin{tabular}{|c|c|c||c|c|c|c|}
\hline
\multicolumn{1}{|c|}{\textbf{Size}} & \multicolumn{1}{c|}{\textbf{Ratio}} & \multicolumn{1}{|c||}{\textbf{Loss}} & \multicolumn{1}{l|}{\textbf{Clean}} & \multicolumn{1}{c|}{\textbf{Blockwise}} & \multicolumn{1}{c|}{\textbf{Random}}  \\ \hline \hline
16 & 25\% & Complete & 4.760 & 5.570 & 4.983  \\ 
16 & 40\% & Complete & 4.872 & 5.594 & 5.079  \\
32 & 25\% & Complete & 4.793 & 5.612 & 5.016  \\
16 & 25\% & Masked & 4.794	& 5.652	& 5.031 \\
\hline
\end{tabular}
\caption{Comparing the impact of mask ratio, mask size, and training target. Values denote RMSE (m).}
\label{tab:ablation_size_ratio_loss_obstacles}
\end{table}

MIMDepth applies the training loss to the complete target image. 
We also compare with applying the loss only on the masked region as used by SimMIM~\cite{xie2022simmim}.  Table~\ref{tab:ablation_size_ratio_loss_obstacles} shows that MIMDepth works better when the loss $\ell_{depth}$ is applied to the entire image. This is due to the training model's focus on learning to predict depth for the complete image, as opposed to the pre-training model's focus on learning representations that only assist downstream tasks at the cost of poor image reconstruction.

\subsection{Impact on Ego-motion estimation Performance}
We additionally evaluate the impact of integrating MIM with depth estimation on the ego-motion estimation predictions. We train MIMDepth on the KITTI odometry split and compare with the baseline MT-SfMLearner in Table~\ref{tab:odom}.  Note that ego-motion predictions on the i.i.d. test set are improved by integrating MIM into the depth network. This can be understood by the use of a common loss for depth and ego-motion networks that is reduced by learning a better context for depth estimation as well as by improving the ego-motion prediction. 

\begin{table}
\centering
\resizebox{\linewidth}{!}{
\begin{tabular}{c|cc|cc|}
\cline{2-5}
 & \multicolumn{2}{c|}{09} & \multicolumn{2}{c|}{10} \\ \hline
\multicolumn{1}{|c|}{\textbf{Method}} & $t_{err} (\%)$ & $r_{err} (^{\circ}/100\text{m})$ & $t_{err} (\%)$ & $r_{err} (^{\circ}/100\text{m})$ \\ \hline \hline
\multicolumn{1}{|c|}{MT-SfMLearner~\cite{visapp22}} & 9.574 & 2.570 & 13.851 & 5.568 \\
\multicolumn{1}{|c|}{MIMDepth (Ours)} & 6.236 & 2.065 & 8.178 & 3.865 \\ \hline
\end{tabular}}
\caption{Comparison of the performance of ego-motion estimation against the baseline in the KITTI odometry split.}
\label{tab:odom}
\end{table}

\section{Conclusion}
We propose a method for adapting masked image modeling (MIM), used until now for self-supervised pre-training of models, into direct training of monocular depth estimation. Our method, MIMDepth, shows improved robustness against noise, blur, weather conditions, digital artifacts, and untargeted and targeted adversarial attacks on depth estimation. While model robustness is often at odds with accuracy, our method is able to maintain comparable depth estimation performance, while also showing improvements in ego-motion estimation. Through ablation studies, we reason about the differences in applying MIM to pre-training and adapting it for depth estimation. We find that blockwise masking (as opposed to random masking) of only the depth estimation network with a relatively lower mask size (than that used for pre-training), and a loss on the complete image (instead of only the masked regions, as used for pre-training) results in more robust depth estimates. We contend that this work could inspire the community to focus on the robustness required in the deployment of depth estimation networks. In the future, such methods could be extended to transformer-based depth estimation with multi-frame input (e.g.~\cite{guizilini2022multi} or overlapping image patches (e.g.~\cite{yang2022depth}). Additionally, MIM could also be adapted for direct training of more downstream tasks.  

\addtolength{\textheight}{-5cm}   

\bibliographystyle{IEEEtran}
\bibliography{IEEEabrv,ref}

\begin{thebibliography}{10}
\providecommand{\url}[1]{#1}
\csname url@rmstyle\endcsname
\providecommand{\newblock}{\relax}
\providecommand{\bibinfo}[2]{#2}
\providecommand\BIBentrySTDinterwordspacing{\spaceskip=0pt\relax}
\providecommand\BIBentryALTinterwordstretchfactor{4}
\providecommand\BIBentryALTinterwordspacing{\spaceskip=\fontdimen2\font plus
\BIBentryALTinterwordstretchfactor\fontdimen3\font minus
  \fontdimen4\font\relax}
\providecommand\BIBforeignlanguage[2]{{%
\expandafter\ifx\csname l@#1\endcsname\relax
\typeout{** WARNING: IEEEtran.bst: No hyphenation pattern has been}%
\typeout{** loaded for the language `#1'. Using the pattern for}%
\typeout{** the default language instead.}%
\else
\language=\csname l@#1\endcsname
\fi
#2}}

\bibitem{schonberger2016structure}
J.~L. Schonberger and J.-M. Frahm, ``Structure-from-motion revisited,'' in
  \emph{Proceedings of the IEEE conference on computer vision and pattern
  recognition}, 2016, pp. 4104--4113.

\bibitem{lee1907big}
J.~Lee, M.~Han, D.~Ko, and I.~Suh, ``From big to small: Multi-scale local
  planar guidance for monocular depth estimation. arxiv 2019,'' \emph{arXiv
  preprint arXiv:1907.10326}, 1907.

\bibitem{yin2019enforcing}
W.~Yin, Y.~Liu, C.~Shen, and Y.~Yan, ``Enforcing geometric constraints of
  virtual normal for depth prediction,'' in \emph{Proceedings of the IEEE
  International Conference on Computer Vision}, 2019, pp. 5684--5693.

\bibitem{chawlavarma2021multimodal}
H.~{Chawla}, A.~{Varma}, E.~{Arani}, and B.~{Zonooz}, ``Multimodal scale
  consistency and awareness for monocular self-supervised depth estimation,''
  in \emph{2021 IEEE International Conference on Robotics and Automation
  (ICRA)}.\hskip 1em plus 0.5em minus 0.4em\relax IEEE, 2021.

\bibitem{guizilini20203d}
V.~Guizilini, R.~Ambrus, S.~Pillai, A.~Raventos, and A.~Gaidon, ``3d packing
  for self-supervised monocular depth estimation,'' in \emph{Proceedings of the
  IEEE/CVF Conference on Computer Vision and Pattern Recognition}, 2020, pp.
  2485--2494.

\bibitem{chawlavarma2022adversarial}
H.~{Chawla}, A.~{Varma}, E.~{Arani}, and B.~{Zonooz}, ``Adversarial attacks on
  monocular pose estimation,'' in \emph{2022 IEEE/RSJ International Conference
  on Intelligent Robotics and Systems (IROS)}.\hskip 1em plus 0.5em minus
  0.4em\relax IEEE (in press), 2022.

\bibitem{visapp22}
A.~Varma., H.~Chawla., B.~Zonooz., and E.~Arani., ``Transformers in
  self-supervised monocular depth estimation with unknown camera intrinsics,''
  in \emph{Proceedings of the 17th International Joint Conference on Computer
  Vision, Imaging and Computer Graphics Theory and Applications - Volume 4:
  VISAPP,}, INSTICC.\hskip 1em plus 0.5em minus 0.4em\relax SciTePress, 2022,
  pp. 758--769.

\bibitem{lee2021realtime}
S.~Lee, C.~Lee, H.~Kim, and H.~J. Kim, ``Realtime object-aware monocular depth
  estimation in onboard systems,'' \emph{International Journal of Control,
  Automation and Systems}, vol.~19, no.~9, pp. 3179--3189, 2021.

\bibitem{bao2021beit}
H.~Bao, L.~Dong, S.~Piao, and F.~Wei, ``Beit: Bert pre-training of image
  transformers,'' in \emph{International Conference on Learning
  Representations}, 2021.

\bibitem{xie2022simmim}
Z.~Xie, Z.~Zhang, Y.~Cao, Y.~Lin, J.~Bao, Z.~Yao, Q.~Dai, and H.~Hu, ``Simmim:
  A simple framework for masked image modeling,'' in \emph{Proceedings of the
  IEEE/CVF Conference on Computer Vision and Pattern Recognition}, 2022, pp.
  9653--9663.

\bibitem{he2022masked}
K.~He, X.~Chen, S.~Xie, Y.~Li, P.~Doll{\'a}r, and R.~Girshick, ``Masked
  autoencoders are scalable vision learners,'' in \emph{Proceedings of the
  IEEE/CVF Conference on Computer Vision and Pattern Recognition}, 2022, pp.
  16\,000--16\,009.

\bibitem{wei2022masked}
C.~Wei, H.~Fan, S.~Xie, C.-Y. Wu, A.~Yuille, and C.~Feichtenhofer, ``Masked
  feature prediction for self-supervised visual pre-training,'' in
  \emph{Proceedings of the IEEE/CVF Conference on Computer Vision and Pattern
  Recognition}, 2022, pp. 14\,668--14\,678.

\bibitem{zhou2017unsupervised}
T.~Zhou, M.~Brown, N.~Snavely, and D.~G. Lowe, ``Unsupervised learning of depth
  and ego-motion from video,'' 2017.

\bibitem{godard2017unsupervised}
C.~Godard, O.~Mac~Aodha, and G.~J. Brostow, ``Unsupervised monocular depth
  estimation with left-right consistency,'' in \emph{Proceedings of the IEEE
  Conference on Computer Vision and Pattern Recognition}, 2017, pp. 270--279.

\bibitem{godard2019digging}
C.~Godard, O.~Mac~Aodha, M.~Firman, and G.~J. Brostow, ``Digging into
  self-supervised monocular depth estimation,'' in \emph{Proceedings of the
  IEEE/CVF International Conference on Computer Vision}, 2019, pp. 3828--3838.

\bibitem{gordon2019depth}
A.~Gordon, H.~Li, R.~Jonschkowski, and A.~Angelova, ``Depth from videos in the
  wild: Unsupervised monocular depth learning from unknown cameras,'' 2019.

\bibitem{klingner2020selfsupervised}
M.~Klingner, J.-A. Termöhlen, J.~Mikolajczyk, and T.~Fingscheidt,
  ``{Self-Supervised Monocular Depth Estimation: Solving the Dynamic Object
  Problem by Semantic Guidance},'' in \emph{ECCV}, 2020.

\bibitem{ranjan2019competitive}
A.~Ranjan, V.~Jampani, L.~Balles, K.~Kim, D.~Sun, J.~Wulff, and M.~J. Black,
  ``Competitive collaboration: Joint unsupervised learning of depth, camera
  motion, optical flow and motion segmentation,'' in \emph{Proceedings of the
  IEEE conference on computer vision and pattern recognition}, 2019, pp.
  12\,240--12\,249.

\bibitem{bian2019unsupervised}
J.~Bian, Z.~Li, N.~Wang, H.~Zhan, C.~Shen, M.-M. Cheng, and I.~Reid,
  ``Unsupervised scale-consistent depth and ego-motion learning from monocular
  video,'' in \emph{Advances in Neural Information Processing Systems}, 2019,
  pp. 35--45.

\bibitem{guizilini2022multi}
V.~Guizilini, R.~Ambruș, D.~Chen, S.~Zakharov, and A.~Gaidon, ``Multi-frame
  self-supervised depth with transformers,'' in \emph{Proceedings of the
  IEEE/CVF Conference on Computer Vision and Pattern Recognition}, 2022, pp.
  160--170.

\bibitem{yang2022depth}
J.~Yang, L.~An, A.~Dixit, J.~Koo, and S.~I. Park, ``Depth estimation with
  simplified transformer,'' in \emph{CVPR Workshops}, 2022.

\bibitem{kenton2019bert}
J.~D. M.-W.~C. Kenton and L.~K. Toutanova, ``Bert: Pre-training of deep
  bidirectional transformers for language understanding,'' in \emph{Proceedings
  of NAACL-HLT}, 2019, pp. 4171--4186.

\bibitem{chen2020generative}
M.~Chen, A.~Radford, R.~Child, J.~Wu, H.~Jun, D.~Luan, and I.~Sutskever,
  ``Generative pretraining from pixels,'' in \emph{International conference on
  machine learning}.\hskip 1em plus 0.5em minus 0.4em\relax PMLR, 2020, pp.
  1691--1703.

\bibitem{dosovitskiy2020vit}
A.~Dosovitskiy, L.~Beyer, A.~Kolesnikov, D.~Weissenborn, X.~Zhai,
  T.~Unterthiner, M.~Dehghani, M.~Minderer, G.~Heigold, S.~Gelly, J.~Uszkoreit,
  and N.~Houlsby, ``An image is worth 16x16 words: Transformers for image
  recognition at scale,'' \emph{ICLR}, 2021.

\bibitem{ramesh2021zero}
A.~Ramesh, M.~Pavlov, G.~Goh, S.~Gray, C.~Voss, A.~Radford, M.~Chen, and
  I.~Sutskever, ``Zero-shot text-to-image generation,'' in \emph{International
  Conference on Machine Learning}.\hskip 1em plus 0.5em minus 0.4em\relax PMLR,
  2021, pp. 8821--8831.

\bibitem{touvron2021training}
H.~Touvron, M.~Cord, M.~Douze, F.~Massa, A.~Sablayrolles, and H.~J{\'e}gou,
  ``Training data-efficient image transformers \& distillation through
  attention,'' in \emph{International Conference on Machine Learning}.\hskip
  1em plus 0.5em minus 0.4em\relax PMLR, 2021, pp. 10\,347--10\,357.

\bibitem{wang2004image}
Z.~Wang, A.~C. Bovik, H.~R. Sheikh, and E.~P. Simoncelli, ``Image quality
  assessment: from error visibility to structural similarity,'' \emph{IEEE
  transactions on image processing}, vol.~13, no.~4, pp. 600--612, 2004.

\bibitem{loshchilov2017decoupled}
I.~Loshchilov and F.~Hutter, ``Decoupled weight decay regularization,''
  \emph{arXiv preprint arXiv:1711.05101}, 2017.

\bibitem{deng2009imagenet}
J.~Deng, W.~Dong, R.~Socher, L.-J. Li, K.~Li, and L.~Fei-Fei, ``Imagenet: A
  large-scale hierarchical image database,'' in \emph{2009 IEEE conference on
  computer vision and pattern recognition}.\hskip 1em plus 0.5em minus
  0.4em\relax Ieee, 2009, pp. 248--255.

\bibitem{geiger2013vision}
A.~Geiger, P.~Lenz, C.~Stiller, and R.~Urtasun, ``Vision meets robotics: The
  kitti dataset,'' \emph{The International Journal of Robotics Research},
  vol.~32, no.~11, pp. 1231--1237, 2013.

\bibitem{kurakin2016adversarial}
A.~Kurakin, I.~Goodfellow, S.~Bengio, \emph{et~al.}, ``Adversarial examples in
  the physical world,'' 2016.

\bibitem{wong2020targeted}
A.~Wong, S.~Cicek, and S.~Soatto, ``Targeted adversarial perturbations for
  monocular depth prediction,'' in \emph{Advances in neural information
  processing systems}, 2020.

\bibitem{tsipras2018robustness}
D.~Tsipras, S.~Santurkar, L.~Engstrom, A.~Turner, and A.~Madry, ``Robustness
  may be at odds with accuracy,'' in \emph{International Conference on Machine
  Learning}, 2019.

\bibitem{eigen2014depth}
D.~Eigen, C.~Puhrsch, and R.~Fergus, ``Depth map prediction from a single image
  using a multi-scale deep network,'' in \emph{Advances in neural information
  processing systems}, 2014, pp. 2366--2374.

\end{thebibliography}

\end{document}